\documentclass[runningheads]{llncs}
\usepackage[T1]{fontenc}
\usepackage{graphicx,verbatim}
\usepackage{algorithm}
\usepackage{algorithmic}

\usepackage{booktabs}       
\usepackage{amsfonts}       
\usepackage{nicefrac}       
\usepackage{xcolor}         
\usepackage[expansion=false]{microtype}
\usepackage{amsmath}
\usepackage{cite}
\usepackage{multirow}
\usepackage{colortbl}
\definecolor{lightpastelpurple}{rgb}{0.69, 0.61, 0.85}
\definecolor{gray}{rgb}{0.75, 0.75, 0.75}
\colorlet{LightGreen}{lightpastelpurple!40}
\colorlet{Gray}{gray!40}

\definecolor{maroon}{cmyk}{0,0.87,0.68,0.32}

\usepackage{tabularx}
\usepackage{tikz}
\usepackage{commath}
\usepackage{lipsum}
\begin{document}
\title{Frequency Adapter with SAM for Generalized Medical Image Segmentation}
%

\author{Phuoc-Nguyen Bui \and
        Van-Nguyen Pham \and \\
        Duc-Tai Le \and
        Junghyun Bum \and
        Hyunseung Choo}
 
\authorrunning{P.-N. Bui et al.}
\institute{Sungkyunkwan University, Korea \\
}

\maketitle              
\begin{abstract}
Medical image segmentation is a critical task in computer-aided diagnosis and treatment planning. However, deep learning models often struggle to generalize across datasets due to domain shifts arising from variations in imaging protocols, scanner types, and patient populations. Traditional domain generalization (DG) methods utilize causal feature learning, adversarial consistency, and style augmentation to improve segmentation robustness. While effective, these approaches rely on explicit feature alignment, adversarial objectives, or handcrafted augmentations, which may not fully exploit the capabilities of foundation models. Recently, the Segment Anything Model (SAM) has demonstrated strong generalization capabilities in segmentation tasks. SAM-based DG methods attempt to improve medical image segmentation. However, these approaches primarily operate in the spatial domain and overlook frequency-based discrepancies that significantly affect model robustness. In this work, we propose Frequency-based Domain Generalization with SAM (FSAM), a novel framework that integrates Low-Rank Adaptation (LoRA) for efficient fine-tuning and a frequency adapter to incorporate frequency-domain representations for single-source domain generalization. FSAM enhances SAM’s segmentation robustness by extracting domain-invariant high-frequency features, mitigating frequency-related domain shifts. Experimental results on fundus and prostate datasets demonstrate that FSAM outperforms existing traditional DG and SAM-based DG approaches in domain generalization. Codes and pre-trained models will be made available on GitHub.


\end{abstract}
\section{Introduction}
Medical image segmentation plays a crucial role in computer-aided diagnosis and treatment planning. However, deep learning models often fail to generalize across datasets from different institutions, imaging devices, and patient populations due to domain shifts \cite{carlucci2019domain,bui2025multi,imans2025unsupervised}. Domain generalization (DG) aims to address this issue by training models to perform well on unseen distributions without requiring access to target domain data during training. Unlike domain adaptation \cite{tzeng2017adversarial}, which assumes access to unlabeled target data for fine-tuning, DG forces models to learn robust and transferable representations that remain effective across diverse environments. Existing DG approaches can be categorized into traditional DG methods and Segment Anything Model (SAM)-based DG methods.

Traditional DG methods enhance segmentation robustness through causal feature learning, adversarial consistency, and style augmentation. CSDG \cite{ouyang2022causality} extracts domain-invariant features to mitigate distribution shifts, while ACL \cite{xu2022adversarial} and contrastive learning \cite{hu2023devil} enforce feature alignment and consistency across different domains. Style-based approaches such as MixStyle \cite{zhou2021domain} and MaxStyle \cite{chen2022maxstyle} augment training by blending styles from multiple domains, encouraging models to learn more generalized representations. However, these approaches rely on explicit feature alignment, adversarial objectives, or handcrafted augmentations, which may not fully leverage foundation models like SAM.

Recently, foundation models such as SAM \cite{kirillov2023segment}, trained on over 1 billion masks, have demonstrated exceptional generalization capabilities across various segmentation tasks. Consequently, researchers have explored adapting SAM for medical image segmentation. Decoupled SAM (DeSAM) \cite{gao2024desam} improves generalization by decoupling domain-specific and domain-agnostic features, while DAPSAM \cite{wei2024prompting} introduces prototype-based prompts to refine SAM’s segmentation in unseen domains. Despite these advancements, existing SAM-based approaches do not explicitly address frequency-based domain shifts, which significantly impact medical imaging due to differences in texture, contrast, and spatial resolution. Some theoretical studies \cite{lin2023deep, li2023frequency} have uncovered that DNNs have preferences for some frequency components in the learning process and indicated that this may affect the robustness of learned features, hence address the domain shift problem.

In this work, we propose Frequency-based Domain Generalization with SAM (FSAM), which integrates Low-Rank Adaptation (LoRA) for efficient fine-tuning of SAM and design a Frequency Adapter to incorporate frequency-domain features to boost the performance for single-source domain generalization. Given a medical image, FSAM extracts features from SAM while leveraging the Frequency Adapter to aggregate high-frequency components, capturing domain-invariant structures. These features are then decoded into segmentation masks using an automated prompt generator \cite{wei2024prompting}. Our main contributions are summarized as follows:

\begin{itemize}
    \item We propose FSAM, a novel frequency-based adaptation of SAM that enhances domain generalization for medical image segmentation.
    \item We design a Frequency Adapter that aggregates high-frequency components, improving robustness against domain shifts.
    \item The experimental results in fundus and prostate datasets demonstrate the superiority of FSAM over traditional DG methods based on SAM.
\end{itemize}

\section{Methodology}
\begin{figure*}[!ht]
  \centering
  \includegraphics[width=0.95\textwidth]{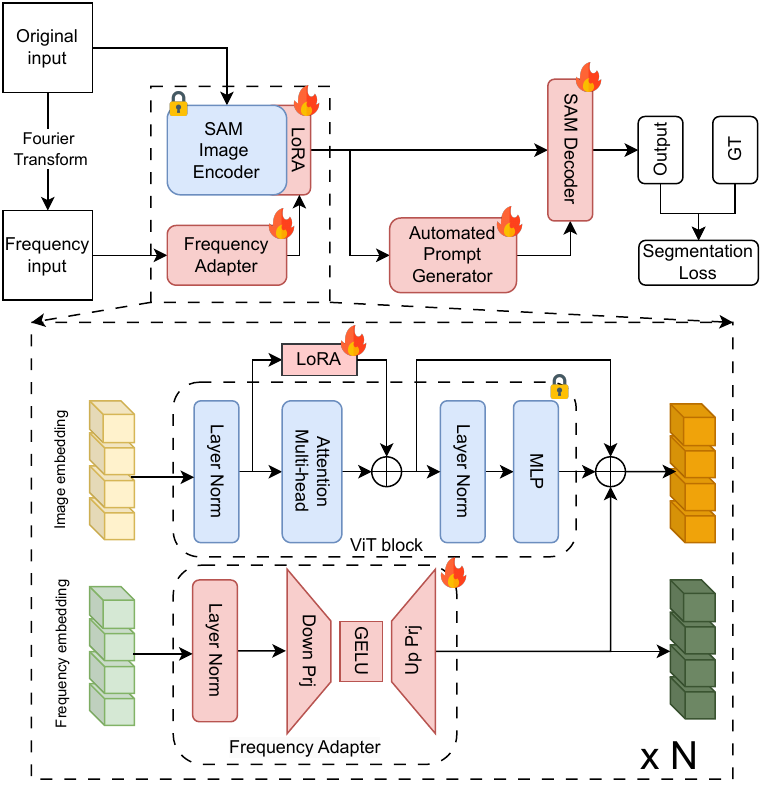}
    \caption{Overview of the proposed frequency-based domain generalization framework with SAM (FSAM). The fire icon represents trainable parameters, while the lock icon indicates frozen parameters retained from the pre-trained model.}
    \label{fig: FSAM}
\end{figure*}

In this section, we first define the single-source domain generalization (SSDG) problem. Then, we elaborate on the proposed FSAM method. Finally, we detail the overall training objective. The SSDG is defined as training on a single source domain and then testing model performance on unseen test domains. SSDG considers a scenario where a model is trained using data from a single labeled source domain $\mathcal{D}_s = \{(x_i^s, y_i^s)\}_{i=1}^{N_s}$, where $x_i^s$ represents medical images and $y_i^s$ denotes the corresponding segmentation masks. The objective is to learn a segmentation model $f_{\theta}$ parameterized by $\theta$ such that it performs well on an unseen target domain $\mathcal{D}_t$, where $\mathcal{D}_s \neq \mathcal{D}_t$ and no target domain samples are available during training. As shown in Figure \ref{fig: FSAM}, we adapt SAM \cite{kirillov2023segment} using LoRA \cite{hu2022lora} and a frequency adapter for single-source domain generalization.

\subsection{FSAM encoder}
Given the input image $I$ of dimension H$\times$W, we obtain the frequency input using Fourier transform. These inputs are processed using SAM's encoder and frequency adapter to leverage the strengths of both spatial and frequency domains.

\textbf{Frequency Adapter} The Discrete Fourier Transform (DFT) is a widely used technique to convert an image from the spatial to the frequency domain. The Fast Fourier Transform (FFT) is practically employed for efficient computation of DFT, the spectrum representation of $f(h,w)$ can be formulated as:
\begin{equation}
    F(u, v) = \sum_{h=0}^{H-1} \sum_{w=0}^{W-1} f(h, w) \cdot e^{-j 2 \pi \left( \frac{uh}{H} + \frac{vw}{W} \right)}
\end{equation}
where $f(h, w)$ represents the pixel intensity at spatial coordinates $(h, w)$ in the image domain, $F(u, v)$ is the corresponding frequency spectrum at frequency coordinates $(u, v)$. The exponential term captures the oscillatory basis functions used to decompose the image into its frequency components. The amplitude and phase components of $F(u,v)$ are derived as $|F(u,v)|$ and $\arg(F(u,v))$, respectively. To efficiently compute the DFT, FFT is applied, reducing the computational complexity from $\mathcal{O}(HW^2)$ in the naive DFT implementation to $\mathcal{O}(HW \log HW)$, making it feasible for medical image processing. 

Next, both the original input and its amplitude component in the frequency domain are processed through the Patch Embedding layer of the ViT architecture \cite{dosovitskiy2020image}, generating the image embedding and frequency embedding, respectively. As illustrated in Fig. \ref{fig: FSAM}, the frequency adapter consists of a linear down-projection layer, followed by a GELU activation function, and a linear up-projection layer. To fully integrate frequency-domain information, we employ $N$ frequency adapters, where $N$ corresponds to the number of ViT blocks in SAM’s encoder.

\textbf{Low-Rank Adaptation (LoRA)} introduces trainable low-rank matrices into the attention layers, enabling efficient adaptation while maintaining the pre-trained knowledge of the foundation model. Specifically, instead of updating the full weight matrix $W \in \mathbb{R}^{d \times d}$ in the self-attention layers of SAM, LoRA parameterizes the weight update as:
\begin{equation}
    W' = W + \Delta W, \quad \text{where} \quad \Delta W = AB
\end{equation}

where $A \in \mathbb{R}^{d \times r}$ and $B \in \mathbb{R}^{r \times d}$ are learnable low-rank matrices, and $r \ll d$ is the rank of the adaptation. This formulation significantly reduces the number of trainable parameters while preserving the expressiveness of the model. LoRA-updated query, key, and value projections are:
\[
\begin{aligned}
W_Q' = W_Q + A_Q B_Q \\
W_K' = W_K + A_K B_K \\
W_V' = W_V + A_V B_V
\end{aligned}
\]

where $W_Q$, $W_K$, and $W_V$ are the frozen projection layers inherited from SAM’s image encoder, while $A_Q$, $B_Q$, $A_V$, and $B_V$ represent the trainable LoRA parameters. By freezing the pre-trained SAM parameters and updating only the LoRA layers, we enable efficient fine-tuning for medical image segmentation while significantly reducing computational overhead. At each stage of SAM’s encoder, the output from the ViT block is combined with the corresponding output from the frequency adapter before being propagated to the next ViT block. The final encoder output is then utilized by SAM’s decoder and the prototype-based prompt generator \cite{wei2024prompting}, which is further detailed in the following section.

\subsection{Automated prompt generator} 
This module, proposed by Wei et al. \cite{wei2024prompting}, employs a memory-based prototype learning approach to enhance domain adaptability. Initially, global average pooling and max pooling are applied to the FSAM encoder output $e_i$ to derive an instance-level prototype $p_i$. The memory bank is structured as a parameterized matrix $M \in \mathbb{R}^{N \times C}$, where $N$ represents the number of stored prototypes, and $C$ denotes their dimensionality. Given a prototype vector $p_i \in \mathbb{R}^{1 \times C}$ extracted from an image embedding, the memory bank retrieves and updates relevant prototypes to generate a domain-adaptive representation $\tilde{p}_i$. The similarity between $p_i$ and stored prototypes in $M$ is computed using cosine similarity:

\begin{equation}
    s_i = \frac{p_i M^T}{\|p_i\| \|M\|}
\end{equation}

The refined prototype $\hat{p}_i$ is then computed via weighted aggregation:

\begin{equation}
    \hat{p}_i = \sum_{j=1}^{N} \alpha_j M_j, \quad \text{where} \quad \alpha_j = \frac{\exp(s_{i,j})}{\sum_{k=1}^{N} \exp(s_{i,k})}
\end{equation}

where $\hat{p}_i$ and $\alpha_j$ represent the refined prototype and the attention weight assigned to each prototype $M_j$. The memory bank dynamically updates throughout training to preserve prototype consistency across different domains. To further refine the embedding, Wei et al. \cite{wei2024prompting} compute the cosine similarity between $\hat{p}_i$ and FSAM’s encoder output, generating an activation map $A_i$ as guidance information. Finally, the refined prototype $\hat{p}_i$, encoder embedding $e_i$, and activation map $A_i$ are concatenated and processed through a $1\times1$ convolutional layer to generate the final prompt for SAM's decoder.

\textbf{Training objective:} Following SAMed \cite{zhang2023customized} and TriD \cite{chen2023treasure}, we employ a hybrid loss function that combines cross-entropy loss and Dice loss to optimize the segmentation model in the source domain. The overall objective function is formulated as:
\begin{equation}
\mathcal{L} = (1 - \lambda) \mathcal{L}_{\text{CE}} + \lambda \mathcal{L}_{\text{Dice}},
\end{equation}

where $\mathcal{L}_{\text{CE}}$ is the standard pixel-wise cross-entropy loss, $\mathcal{L}_{\text{Dice}}$ is the Dice loss for region-based supervision, and $\lambda$ is a weighting factor that balances the contribution of both loss terms. 

\section{Experiments}
In this section, we conduct comprehensive experiments to assess the effectiveness of FSAM in domain generalization for medical image segmentation. We first outline the experimental setup, including datasets, evaluation metrics, and training configurations. Next, we provide a quantitative comparison between FSAM and prior approaches, covering both traditional and SAM-based techniques.

\subsection{Experimental setup and implementation details}
\textbf{Dataset:} To evaluate the performance of the proposed FSAM, we conducted assessments in two cross-domain settings: 

The first dataset is RIGA+ dataset \cite{decenciere2014feedback, hu2022domain}, which is a multi-domain benchmark for joint optic cup (OC) and optic disc (OD) segmentation, comprising annotated fundus images from five distinct domains: BinRushed, Magrabia, Messidor Base 1, Messidor Base 2, and Messidor Base 3. For our segmentation experiments, we use BinRushed and Magrabia as the source domains for training, while the remaining three domains are treated as target domains to evaluate cross-domain generalization. We adopt the official dataset split. To ensure consistency across different domains, all images are resized to 512$\times$512. Model performance is evaluated using the Dice Similarity Coefficient (DSC) as the primary metric.
\[ DSC = \frac{2\big|GT \cap PR\big|}{\big|GT\big| + \big|PR\big|} =\frac{2TP}{2TP + FP + FN} \]

The second dataset is the prostate dataset \cite{liu2020shape}, which consists of 116 cases spanning six different domains: A (RUNMC), B (BMC), C (I2CVB), D (UCL), E (BIDMC), and F (HK). All slices are resized to 384$\times$384 with standardized voxel spacing for consistency. We follow the preprocessing pipeline of MaxStyle \cite{chen2022maxstyle}. For each experiment, one domain is designated as the source domain, where the data is split into training (90\%) and in-domain validation (10\%). The trained model is then evaluated on all other unseen domains to assess cross-domain generalization. The DSC is also employed as the primary performance metric.

\textbf{Implementation details}: We configure the LoRA rank to 4, balancing efficiency and performance. The frequency adapters incorporate amplitude information to enhance feature extraction. Training is conducted using the ViT-B variant of SAM, with an initial learning rate of 5e-4 and weight decay of 0.1 for the AdamW optimizer \cite{loshchilov2017decoupled}. Following SAMed \cite{zhang2023customized}, we adopt a warm-up strategy, setting warm-up periods to 250 for the prostate dataset and 25 for RIGA+, considering their differing data configurations. Training is early stopped at 160 epochs, with a maximum of 200 epochs, and the loss weighting factor is set to 0.8. The code and pre-trained models will be released on GitHub for reproducibility.

\subsection{Performance evaluation}

\textbf{RIGA+ dataset:} As presented in Table \ref{tab:2}, our proposed FSAM demonstrates superior performance compared with both traditional and SAM-based methods. When using Magrabia as the source domain, FSAM outperforms the CNN-based state-of-the-art CCSDG, achieving a 1.41\% improvement in optic disc segmentation (96.39\% vs. 94.98\%) and a 3.10\% gain in optic cup segmentation (88.63\% vs. 85.53\%). Additionally, FSAM attains competitive or superior performance compared to other SAM-based methods, further validating its effectiveness.

\begin{table*}[!h]
  \caption{Segmentation performance on the multi-site RIGA+ fundus dataset. The best and second-best results are highlighted in \textbf{bold} and \underline{underlined}, respectively. The table is divided into two parts: the upper part corresponds to experiments using BinRushed as the source domain, while the lower part represents results obtained with Magrabia as the source domain.
} 
  \centering
  \setlength{\tabcolsep}{4pt}
  \resizebox{\linewidth}{!}{
  \begin{tabular}{lcc|cc|cc|cc|cc}
    \toprule
    \multirow{2}{*}{Method} & \multirow{2}{*}{Year} & \multirow{2}{*}{Backbone} & \multicolumn{2}{c|}{Messidor Base 1} & \multicolumn{2}{c|}{Messidor Base 2} & \multicolumn{2}{c|}{Messidor Base 3} & \multicolumn{2}{c}{Average} \\
    & & & OD & OC & OD & OC & OD & OC & OD & OC \\
    \midrule
    \multicolumn{11}{c}{\textbf{BinRushed}} \\
    \midrule
    CSDG \cite{ouyang2022causality} & 2022 & \multirow{6}{*}{UNet} & 93.56 & 81.00 & 94.38 & 83.79 & 93.87 & 83.75 & 93.93 & 82.85 \\
    ADS \cite{xu2022adversarial} & 2022 &  & 94.07 & 79.60 & 94.29 & 81.17 & 93.64 & 81.08 & 94.00 & 80.62 \\
    MaxStyle \cite{chen2022maxstyle} & 2022 & & 94.28 & 82.61 & 86.65 & 74.71 & 92.36 & 82.33 & 91.09 & 79.88 \\
    D-Norm \cite{zhou2022generalizable} & 2022 & & 94.57 & 81.81 & 93.67 & 79.16 & 94.82 & 83.67 & 94.35 & 81.55 \\
    SLAug \cite{su2023rethinking}& 2023 & & 95.28 & 83.31 & 95.49 & 81.36 & 95.57 & 84.38 & 95.45 & 83.02 \\
    CCSDG \cite{hu2023devil} & 2023 & & 95.73 & 86.13 & 95.73 & 86.28 & 95.45 & 86.77 & 95.64 & 86.57 \\
    \midrule
    SAMed \cite{zhang2023customized} & 2023 & \multirow{5}{*}{ViT} & 95.28 & 84.24 & 94.11 & 80.21 & 94.84 & 82.60 & 94.74 & 82.35 \\
    DeSAM \cite{gao2024desam} [box] & 2024 & & 89.33 & 79.68 & 93.44 & 82.97 & 91.51 & 82.70 & 91.42 & 81.78 \\
    DeSAM \cite{gao2024desam} [points] & 2024 &  & 91.79 & 80.87 & 92.57 & 82.95 & 93.66 & 84.19 & 92.67 & 82.67 \\
    DAPSAM \cite{wei2024prompting}& 2024 & & \textbf{96.34} & \textbf{88.24} & \underline{96.10} & \textbf{86.31} & \textbf{96.34} & \textbf{88.77} & \textbf{96.26} & \textbf{87.77}\\
    \rowcolor{Gray} \textbf{FSAM} & 2025 & & \underline{96.13} & \underline{88.12} & \textbf{96.29} & \underline{86.20} & \underline{96.32} & \underline{88.28} & \underline{96.25} & \underline{87.53} \\
    \midrule
    \multicolumn{11}{c}{\textbf{Magrabia}} \\
    \midrule
    CSDG \cite{ouyang2022causality} & 2022 & \multirow{6}{*}{UNet} & 89.67 & 75.39 & 87.97 & 76.44 & 89.91 & 81.35 & 89.18 & 77.73 \\
    ADS \cite{xu2022adversarial} & 2022 &  & 90.75 & 77.78 & 90.37 & 79.60 & 90.34 & 79.99 & 90.48 & 79.12 \\
    MaxStyle \cite{chen2022maxstyle} & 2022 & & 91.63 & 78.74 & 90.61 & 80.12 & 91.22 & 81.90 & 91.15 & 80.25 \\
    D-Norm \cite{zhou2022generalizable} & 2022 & & 92.35 & 79.02 & 91.23 & 80.06 & 92.09 & 79.87 & 91.89 & 79.65 \\
    SLAug \cite{su2023rethinking} & 2023 & & 93.08 & 80.70 & 92.70 & 80.15 & 92.23 & 80.89 & 92.67 & 80.58 \\
    CCSDG \cite{hu2023devil} & 2023 & & 94.78 & 84.94 & 95.16 & 85.68 & 95.00 & 85.98 & 94.98 & 85.53 \\
    \midrule
    SAMed \cite{zhang2023customized} & 2023 & \multirow{5}{*}{ViT} & 95.41 & 85.26 & 95.36 & 84.25 & 95.38 & 84.76 & 95.38 & 84.76 \\
    DeSAM \cite{gao2024desam} [box] & 2024 & & 82.45 & 69.66 & 84.97 & 75.75 & 83.86 & 74.74 & 83.76 & 73.38 \\
    DeSAM \cite{gao2024desam} [points] & 2024 &  & 81.39 & 67.88 & 83.95 & 76.33 & 79.99 & 73.05 & 84.50 & 72.42 \\
    DAPSAM \cite{wei2024prompting} & 2024 & & \underline{96.22} & \underline{86.74} & \underline{96.32} & \underline{89.59} & \underline{96.35} & \underline{88.12} & \underline{96.30} & \underline{88.15} \\
    \rowcolor{Gray} \textbf{FSAM} & 2025 & & \textbf{96.28} & \textbf{87.61} & \textbf{96.41} & \textbf{90.07} & \textbf{96.48} & \textbf{88.20} & \textbf{96.39} & \textbf{88.63} \\
    \bottomrule
  \end{tabular}
  }
  \label{tab:2}
\end{table*}

\textbf{Prostate dataset:} Compared to the traditional CNN-based U-Net architecture \cite{ronneberger2015u}, ViT-based methods built on SAM demonstrate superior segmentation performance, as shown in Table \ref{tab:1}. Our proposed method not only surpasses the best-performing CNN-based approach but also outperforms recent SAM-based methods on the prostate dataset. Specifically, compared to CSDG, the top-performing CNN-based method, FSAM achieves a significant 12.68\% improvement, highlighting the effectiveness of incorporating frequency-aware adaptation. Furthermore, our method outperforms the recently proposed SAM-based approach, DAPSAM \cite{wei2024prompting}, achieving an additional 1.43\% performance gain, demonstrating the advantage of integrating LoRA fine-tuning and frequency-domain representations for robust medical image segmentation.

\begin{table*}[!h]
  \caption{Segmentation performance on multi-site prostate dataset. The best and second-best results are highlighted in \textbf{bold} and \underline{underline}, respectively. Each column represents leave-one-out results for the model trained on the corresponding domain while testing on the other domains.}
  \centering
  \setlength{\tabcolsep}{4pt}
  \resizebox{\linewidth}{!}{
  \begin{tabular}{lcc|cccccc|c}
    \toprule
    Method & Year & Backbone & A & B & C & D & E & F & Average \\
    \midrule
    In-domain & - & UNet & 85.38 & 83.68 & 82.15 & 85.21 & 87.04 & 84.29 & 84.63 \\
    \midrule
    AdvBias \cite{chen2020realistic}& 2020 & \multirow{5}{*}{UNet} & 77.45 & 62.12 & 51.09 & 70.20 & 51.12 & 50.69 & 60.45 \\
    RandConv \cite{xu2020robust} & 2020 &  & 75.52 & 57.23 & 44.21 & 61.27 & 49.98 & 54.21 & 57.07 \\
    MixStyle \cite{zhou2021domain} & 2021 & & 73.04 & 59.29 & 43.00 & 62.17 & 53.12 & 50.03 & 56.78 \\
    MaxStyle \cite{chen2022maxstyle} & 2022 & & 81.25 & 70.27 & 62.09 & 58.18 & 70.04 & 67.77 & 68.27 \\
    CSDG \cite{ouyang2022causality} & 2022 & & 80.72 & 68.00 & 59.78 & 72.40 & 68.67 & 70.78 & 70.06 \\
    \midrule
    SAMed \cite{zhang2023customized}& 2023 & \multirow{6}{*}{ViT} & 80.42 & \underline{81.44} & 66.75 & 82.09 & 80.19 & 80.17 & 78.51 \\
    DeSAM \cite{gao2024desam} [box] & 2024 & & 82.30 & 78.06 & 66.65 & 82.87 & 77.58 & 79.05 & 77.75 \\
    DeSAM \cite{gao2024desam} [points] & 2024 &  & 82.80 & 80.61 & 64.77 & 83.41 & 80.36 & \underline{82.17} & 79.02 \\
    MedSAM \cite{ma2024segment}& 2024 & & 72.32 & 73.31 & 61.53 & 64.46 & 68.89 & 61.39 & 66.98 \\
    DAPSAM \cite{wei2024prompting}& 2024 & & \textbf{86.34} & 81.05 & \underline{70.81} & \underline{85.28} & \underline{82.91} & 81.48 & \underline{81.31} \\
    \rowcolor{Gray}\textbf{FSAM} & 2025 & & \underline{86.16} & \textbf{82.52} & \textbf{74.98} & \textbf{86.44} & \textbf{83.14} & \textbf{83.19} & \textbf{82.74} \\
    \bottomrule
  \end{tabular}
  }
  \label{tab:1}
\end{table*}


\section{Conclusion}
In this work, we propose Frequency-based Domain Generalization with Segment Anything Model (FSAM), a novel framework designed to enhance the generalization capability of SAM for medical image segmentation. FSAM leverages Low-Rank Adaptation (LoRA) for efficient fine-tuning and integrates a Frequency Adapter to extract domain-invariant frequency features, effectively mitigating domain shifts caused by variations in imaging protocols, scanner types, and patient populations. By incorporating frequency-domain representations, FSAM captures essential structural and texture-based information that remains robust across different domains. Comprehensive experiments on fundus and prostate datasets demonstrate that FSAM consistently outperforms both traditional domain generalization methods and SAM-based DG approaches, achieving superior segmentation accuracy on unseen domains. These results highlight the efficacy of combining frequency-aware adaptation with LoRA-based fine-tuning in foundation models to improve cross-domain robustness.


\subsubsection{\discintname}
The authors have no competing interests to declare that are relevant to the content of this article.

\bibliographystyle{splncs04}
\bibliography{mybibliography}
\end{document}